# G2R Bound: A Generalization Bound for Supervised Learning from GAN-Synthetic Data

**Fu-Chieh Chang**
HTC Research & Healthcare
(DeepQ)
Taipei, Taiwan
`mark.fc_chang@htc.com`

**Hao-Jen Wang**
HTC Research & Healthcare
(DeepQ)
Taipei, Taiwan
`hjwang0086@gmail.com`

**Chun-Nan Chou**
HTC Research & Healthcare
(DeepQ)
Taipei, Taiwan
`jason.cn_chou@htc.com`

**Edward Y. Chang**
Stanford University
Stanford, CA
`eyuchang@gmail.com`

## Abstract

Performing supervised learning from the data synthesized by using Generative Adversarial Networks (GANs), dubbed *GAN-synthetic data*, has two important applications. First, GANs may generate more labeled training data, which may help improve classification accuracy. Second, in scenarios where real data cannot be released outside certain premises for privacy and/or security reasons, using GAN-synthetic data to conduct training is a plausible alternative. This paper proposes a generalization bound to guarantee the generalization capability of a classifier learning from GAN-synthetic data. This generalization bound helps developers gauge the generalization gap between learning from synthetic data and testing on real data, and can therefore provide the clues to improve the generalization capability.

## 1   Introduction

Recent progress in generative adversarial networks (GANs) [11] shows promising results for learning data distributions. For example, the works of [5, 14] propose methods to generate high resolution images with high fidelity. Since GAN-synthetic examples can be nearly indistinguishable from real data, is it possible for a classifier trained on synthetic data to have good generalization capability (i.e., achieve good accuracy on real data)? Several prior works [12, 26, 6, 27] have demonstrated that learning from GAN-synthetic data can be comparable in accuracy to learning from real data. These works, however, report only experimental results, and lack of theoretical explanations for their supposed generalization capability. Hence, their results may be a consequence of chance and not of a theoretical trend. Although GANs can synthesize a large amount of labeled training data, if the quality of synthetic data is not good (i.e., the features of synthetic data are not similar to the features of real data, or the labels of synthetic data are inconsistent with the labels of real data), the classifier learning from synthetic data may not be able to generalize its knowledge well to real data. Without any theoretical guarantees, we cannot estimate whether the quality of synthetic data is good enough to achieve good generalization capability.

In summary, this work makes the following contributions:

1. We propose a new generalization bound to theoretically guarantee that a classifier trained on GAN-synthetic data can be generalized to real data.

2. We propose a method to estimate the actual value of this generalization bound based on finite samples from real datasets and GAN-synthetic datasets.

3. Based on this generalization bound, we uncover the factors influencing the generalization gap between learning from synthetic data and testing on real data. This gap is due to:

    (a) The discrepancy between the synthetic features and the real features, and

    (b) The inconsistency between the labeling function of synthetic data and the labeling function of real data.

4. Our experiments attest that mitigating the generalization gap improves generalization capability, which is consistent with our proposed theorem.

## 2 Related Works

### 2.1 Supervised Learning from GAN-Synthetic Data

Recent progress in generative models, in particular generative adversarial networks (GANs) [11], has resulted in the possibility of synthesizing more training examples for improving classification accuracy and/or preserving data privacy. The works of [30, 26] show that synthesizing more training examples can improve classification accuracy. Several works [12, 26, 6] show through empirical studies that a classifier trained on GAN-synthetic examples can be generalized to classify real-world data to achieve privacy preservation. The work of [27] shows that the generalization capability of learning from synthetic data serves as a good criterion to evaluate the quality of synthetic data. However, their studies only rely on empirical evidence and lack theoretical forms of explanation. Although the work [30] provides a theoretical proof of convergence of the loss function, we argue that convergence of the loss function evaluated on synthetic data cannot be related to the generalization capability on real data. Hence, none of these prior works have a theoretical guarantee for their generalization capability on real data.

Another approach to adopt GAN (or adversarial training) for supervised learning tasks is to train a GAN as a transformer between different domains, or to apply adversarial loss to reduce the gap between two domains. This method can be applied to solve the problem of domain adaptation [9, 29, 28] as well as privacy issues that arises from using real data [7]. The work of [9] shows that adversarial training can improve the generalization bound for domain adaptation tasks. However, in their scenarios, there is no new training example synthesized. Our proposed generalization bound can be applied to the scenario of training on new GAN-synthetic examples.

### 2.2 Generalization Bound

The generalization capability (i.e., the model in which learning from training data can be generalized to testing data) of a supervised machine learning model is guaranteed by the VC generalization bound [1]. This bound guarantees the difference between the actual risk (i.e., the error on the actual data distribution) and the empirical risk (i.e., the error on finite training examples) is not larger than certain value. Hence, a model trained on finite training examples can be generalized to unseen examples in the same distribution. Besides VC generalization bound, there are other generalization bounds for various machine methods, such as multitask learning [20, 3, 21], transfer learning [8, 22, 16], and domain adaptations [19, 4, 10]. Although the generalization bound proposed by the work of [4] seems to be able to adapt to the scenario of learning from GAN-synthetic data, we found that its bound suffers from over-estimation in our problem setting of generalizing supervised learning from GAN-synthetic data to real data.

## 3 Supervised Learning from GAN-Synthetic Data

In this section, we formalize the method of supervised learning from GAN-synthetic data and establish some requisite notations to develop the generalization bound. We first review some background of GANs, followed by two improved versions of GANs. Subsequently, we elaborate the details of the method and specify the bound that we will derive in the next section.

## 3.1 Generative Adversarial Networks

GANs [11] were recently proposed as a promising framework for estimating generative models via adversarial training. This framework consists of a generator and a discriminator. The generator takes a noise vector $z$ sampling from a prior distribution $p(z)$ as the input and outputs an image $x_g$ [1]. The input of the discriminator is either a real image $x_r$ from the distribution of real images $X_r$ or an image $x_g$ from the generator distribution $X_g$. The output of the discriminator $s$ predicts that its input $x$ is an image from $X_r$ or $X_g$, which we denote as $s = 1$ or $s = 0$, respectively. To maximize the accuracy of the predicted sources, the discriminator is trained to maximize the following log-likelihood:

$$L_s = \mathbb{E}_{x_r \sim X_r}[\log(s)] + \mathbb{E}_{z \sim p(z)}[\log(1-s))] . \quad (1)$$

Meanwhile, the generator is trained to minimize the same equation above. Hence, the loss functions of the discriminator and generator are

$$L_{dis} = -L_s \text{ and } L_{gen} = L_s . \quad (2)$$

## 3.2 Conditional GANs and Auxiliary-Classifier GANs

In addition to $X_g$, GANs can be further extended to simultaneously generate some extra information such as class labels. Conditional GANs (cGANs) [23] provide a way to generate a labeling function $f_g$, which maps the feature space to the label space to generate a synthetic labeled dataset $D_g = \langle X_g, f_g \rangle$. The labeling function $f_g$ is analogous to the labeling function $f_r$ on a real labeled $D_r$, where $D_r = \langle X_r, f_r \rangle$. More specifically, every $x_g$ is generated under the condition of a label $y_g$. Therefore, every $x_g$ has its corresponding $y_g$, and the labeling function $f_g$ can be obtained via a look-up table containing every pair of $(x_g, y_g)$.

Among the family of cGANs, Auxiliary-Classifier GANs (AC-GANs) [25] outperform the other approaches and demonstrate the best results. The discriminator of AC-GANs shares its feature extractor with an additional classifier, called *auxiliary classifier*. For every input image, this auxiliary classifier yields the class label $y$. For example, in the handwritten digit recognition problem, $y$ can range from 0 to 9. The loss function of AC-GANs thus comprises the log-likelihood functions of the correct source and class,

$$\begin{aligned} L_s &= \mathbb{E}_{x_r \sim X_r}[\log(s)] + \mathbb{E}_{x_g \sim X_g}[\log(1-s)] \text{ and} \\ L_c &= \mathbb{E}_{x_r \sim X_r}[y_r \log(y)] + \mathbb{E}_{x_g \sim X_g}[y_g \log(y)] . \end{aligned} \quad (3)$$

The additional term $L_c$ in AC-GANs forces a real image $x_r$ and a synthetic image $x_g$ to be classified as $y_r$ and $y_g$, respectively. Accordingly, their discriminators aim to maximize $L_s$ and $L_c$, whereas their generators aim to minimize $L_s$ but maximize $L_c$. The resultant loss functions $L_{dis}$ and $L_{gen}$ are

$$L_{dis} = -L_s - \alpha L_c \text{ and } L_{gen} = L_s - \alpha L_c , \quad (4)$$

where $\alpha$ is the hyper-parameter of $L_c$.

## 3.3 Detailed Method with Notations of Generalization Bounds

Before elaborating the details of supervised learning from GAN-synthetic data, we first introduce the notations and concepts about generalization bound. A labeled dataset $D = \langle X, f \rangle$ can also be regarded as a set of tuples $D = \{(x, y) : y = f(x), x \sim X\}$, where $X$ is the distribution of input features, and $f$ is a labeling function. The notation $(x, y)$ denotes a sample pair from $D$. A training dataset $\hat{D} = \langle \hat{X}, f \rangle$ consists of finite $n$ sample pairs from $D$, and a testing dataset $\tilde{D} = \langle \tilde{X}, f \rangle$ is composed of the other finite $m$ sample pairs from $D$. The symbol $h$ indicates a hypothesis that is a function mapping the feature space to the label space. The notation $\mathbb{I}[\cdot]$ is an indicator function that is evaluated to 1 if its argument is true. We define the actual risk as $\epsilon(h, f) = \mathbb{E}_{x \sim X}[\mathbb{I}[f(x) \neq h(x)]]$ and the empirical risk as $\hat{\epsilon}(h, f) = \mathbb{E}_{x \sim \hat{X}}[\mathbb{I}[f(x) \neq h(x)]] = \frac{1}{n} \sum_{i=1}^{n} \mathbb{I}[f(x_i) \neq h(x_i)]$. The testing error is defined as $\tilde{\epsilon}(h, f) = \mathbb{E}_{x \sim \tilde{X}}[\mathbb{I}[f(x) \neq h(x)]] = \frac{1}{m} \sum_{i=1}^{m} \mathbb{I}[f(x_i) \neq h(x_i)]$. If the dataset notation has a subscript, e.g., $D_s = \langle X_s, f_s \rangle$, its three kinds of risks are denoted as $\epsilon_s(h, f_s)$, $\hat{\epsilon}_s(h, f_s)$, and $\tilde{\epsilon}_s(h, f_s)$.

---
[1] In the first work of GANs, the output is an image. With the advance of research on GANs, the output can be text [13] or other modalities of data. However, we focus on discussing the image scenario throughout this paper.

Given a labeled dataset $D = \langle X, f \rangle$, and a hypothesis space $H$, a machine learning algorithm aims to find a hypothesis $h$ that is close to $f$ (i.e., the values of $\epsilon(h, f)$ is small). However, machine learning algorithm only has limited training data $\hat{D}$, finite samples from $D$, and can only search the hypothesis $h \in H$ based on the value of $\hat{\epsilon}(h, f)$. The VC generalization bound can guarantee a small value of $\epsilon(h, f)$ given a small value of $\hat{\epsilon}(h, f)$, and hence ensures that $h$ can be generalized to $D$. With the probability $1 - \delta$, the following inequalities [1] is satisfied:

$$\epsilon(h, f) \leq \hat{\epsilon}(h, f) + \sqrt{\frac{8}{n}\log(\frac{4(2n)^d}{\delta})}, \quad (5)$$

where $d$ is the VC-dimension [31] of the hypothesis space $H$.

The detailed steps of supervised learning from GAN-synthetic data are depicted as follows:

Step 1. A dataset $D_r = \langle X_r, f_r \rangle$ is given. A cGAN (or AC-GAN) is trained on $\hat{D}_r$.

Step 2. When the training is complete, the trained cGAN (or AC-GAN) synthesizes a new dataset $D_g = \langle X_g, f_g \rangle$ by the method specified in Sec.3.2.

Step 3. A classifier is trained on $\hat{D}_g$. The training algorithm selects a hypothesis $h \in H$ with empirical risk $\hat{\epsilon}_g(h, f_g)$.

Step 4. This hypothesis $h$ is evaluated on $D_r$, and its actual risk is $\epsilon_r(h, f_r)$.

Assume that the $H$ in Step 3. has VC-dimension $d$, and $h$ is selected after training. If the number of training examples $n$ is large enough, the difference between $\hat{\epsilon}_g(h, f_g)$ and $\epsilon_g(h, f_g)$ can be small enough according to Eq. (5), and $h$ can be generalized to $D_g$. However, can $h$ be generalized to $D_r$? If yes, under what conditions can a classifier trained on $\hat{D}_g$ be expected to perform well on $D_r$? We present a bound to guarantee this generalization in the next section.

## 4 Bounds for Supervised Learning from GAN-Synthetic Data

Now, we proceed to develop bounds for supervised learning from GAN-synthetic data, i.e., bounds on the real data generalization performance of a classifier trained on GAN-synthetic data. We first review a domain adaptation (DA) bound introduced in the previous work [4] and adapt the DA bound for our problem setting. Subsequently, we show that the DA bound is prone to over-estimate the bounds in our setting and thus propose a GAN-to-real bound, which is dubbed G2R bound. Finally, we explain how to practically estimate G2R and DA bounds.

### 4.1 Domain Adaptation Bound

Considering the domain adaptation setting, the actual risk $\epsilon_t(h, f_t)$ is evaluated on a target dataset $D_t = \langle X_t, f_t \rangle$, and the training dataset $\hat{D}_s$ is sampled from another source dataset $D_s = \langle X_s, f_s \rangle$. Since $\hat{D}_s$ is not sampled from $D_t$, Eq. (5) cannot be directly applied to this setting. To overcome this issue, the work of [4] proposes a generalization bound, which we call the domain adaptation (DA) bound, to establish the relationship between $\epsilon_s(h, f_s)$ and $\epsilon_t(h, f_t)$. There are two key ideas behind the DA bound. First, DA considers the distance between $X_s$ and $X_t$. If the distance is small, it's very likely that the generalization capability is better. Based on the disagreements between the two hypotheses in $H$, the work of [4] proposes a method to measure this distance, which is called $d_{H \Delta H}$-distance and defined by

$$d_{H \Delta H}(X_s, X_t) = 2 \sup_{g \in H \Delta H} \left| \mathbb{P}_{x \sim X_s}[g(x) = 1] - \mathbb{P}_{x \sim X_t}[g(x) = 1] \right|, \quad (6)$$

where $H \Delta H$ is a hypothesis set of the disagreements between hypotheses $h'$ and $h"$ and defined by

$$g \in H \Delta H \iff g(x) = \mathbb{I}[h'(x) \neq h"(x)] \text{ for some } h', h" \in H. \quad (7)$$

Second, the work of [4] considers a hypothesis $h' \in H$ jointly minimizes the risk on both $D_s$ and $D_t$. The optimal hypothesis $h^*$ is defined as the best hypothesis to minimize the following risk:

$$h^* = arg \min_{h' \in H} \epsilon_s(h', f_s) + \epsilon_t(h', f_t), \quad (8)$$

and the combined error of $h^*$ is

$$\lambda = \epsilon_s(h^*, f_s) + \epsilon_t(h^*, f_t) \,. \tag{9}$$

Based on these two ideas, the work of [4] proposes the following DA bound [2]:

$$\epsilon_t(h, f_t) \leq \epsilon_s(h, f_s) + \lambda + \frac{1}{2} d_{H \Delta H}(X_s, X_t) \,. \tag{10}$$

In our problem setting, a classifier is trained on GAN-synthetic (source) data, and its actual risk is evaluated on real (target) data. Next, we will show theoretically and empirically that the DA bound is not suitable in our GAN-synthetic data generalization setting.

### 4.2 Over-Estimation of Domain Adaptation Bound

When the DA bound in Eq. (10) is applied to the scenario mentioned in Sec.3.3, the bound becomes

$$\epsilon_r(h, f_r) \leq \epsilon_g(h, f_g) + \lambda + \frac{1}{2} d_{H \Delta H}(X_g, X_r) \,. \tag{11}$$

However, adapting this bound for our scenario turns out to be overestimated, and the over-estimation stems from the distance $d_{H \Delta H}(X_g, X_r)$. As mentioned in the Sec 2.1 of [2], if $X_r$ is a real feature distribution, and $X_g$ is a GAN-synthetic feature distribution, there exists a hypothesis $h'$ such that

$$\mathbb{P}_{x \sim X_r}[h'(x) = 1] = 1 \text{ , and } \mathbb{P}_{x \sim X_g}[h'(x) = 1] = 0 \,. \tag{12}$$

Hence, the following equation is satisfied:

$$\big| \mathbb{P}_{x \sim X_r}[h'(x) = 1] - \mathbb{P}_{x \sim X_g}[h'(x) = 1] \big| = 1 \,. \tag{13}$$

The evaluation of $d_{H \Delta H}(X_r, X_g)$ involves searching a hypothesis $h' \in H \Delta H$ to maximize $|\mathbb{P}_{x \sim X_r}[h'(x) = 1] - \mathbb{P}_{x \sim X_g}[h'(x) = 1]|$. If there exists an $h' \in H \Delta H$ satisfying Eq. (13), $d_{H \Delta H}(X_r, X_g)$ will be equal to 2, and the distance between $X_r$ and $X_g$ may be overestimated. In our scenario, $d_{H \Delta H}(X_r, X_g)$ will be overestimated when the following two conditions are satisfied: First, in Sec.3.3, the hypothesis $h$ in Step 4. is selected from a hypothesis space $H$ containing $h'$ satisfying Eq. (13). Second, $H$ contains a constant function $f(x) = 0$, and it is clear that $H \subset H \Delta H$. These two conditions are usually satisfied when the architecture of the classifier has enough expressive power, e.g., a deep convolutional net.

### 4.3 GAN-to-Real Bound

To alleviate this over-estimation arose from $d_{H \Delta H}(X_g, X_r)$, we propose a new measurement of the distance between $X_g$ and $X_r$, called $h \Delta h^*$-distance.

**Definition 4.1.** *Given two feature distributions $X_g$ and $X_r$, and two hypotheses $h$ and $h^*$, the $h \Delta h^*$-distance between $X_g$ and $X_r$ is defined as*

$$d_{h \Delta h^*}(X_g, X_r) = 2 \big| \mathbb{P}_{x \sim X_g}[h(x) \neq h^*(x)] - \mathbb{P}_{x \sim X_r}[h(x) \neq h^*(x)] \big| \,. \tag{14}$$

Our proposed $h \Delta h^*$-distance enjoys the property of a pseudo-metric[3]. The over-estimation problem in Sec.4.2 is also alleviated since the evaluation of $h \Delta h^*$-distance does not involve searching a hypothesis $h \in H \Delta H$ to maximize the distance. Moreover, $d_{h \Delta h^*}(X_g, X_r)$ is always smaller than or equal to $d_{H \Delta H}(X_g, X_r)$ when $h, h^* \in H$ due to the following fact:

$$\begin{aligned} d_{h \Delta h^*}(X_g, X_r) &= 2 \big| \mathbb{P}_{x \sim X_g}[h(x) \neq h^*(x)] - \mathbb{P}_{x \sim X_r}[h(x) \neq h^*(x)] \big| \\ &\leq 2 \sup_{h', h'' \in H} \big| \mathbb{P}_{x \sim X_g}[h'(x) \neq h''(x)] - \mathbb{P}_{x \sim X_r}[h'(x) \neq h''(x)] \big| \\ &= d_{H \Delta H}(X_g, X_r) \,. \end{aligned} \tag{15}$$

Based on the aforementioned reasons, we propose a new generalization bound called GAN-to-Real (G2R) bound by replacing the term $\frac{1}{2} d_{H \Delta H}(X_g, X_r)$ with $\frac{1}{2} d_{h \Delta h^*}(X_g, X_r)$ in Eq. (11). The detailed proof in Supplementary Materials S1.1 shows the validity of this inequality.

---

[2] Please see [4] for the detailed proofs

[3] A pseudo-metric $d$ is a metric that satisfies the condition $x = y \Rightarrow d(x, y) = 0$ but does not satisfy the condition $d(x, y) = 0 \Rightarrow x = y$, and it is easy to verify that $h \Delta h^*$-distance is a pseudo-metric.

**Theorem 4.1.** *Given a real dataset $D_r = \langle X_r, f_r \rangle$, a cGAN is trained on $\hat{D}_r$ based on the steps described in Sec.3.3, this cGAN generates a synthetic dataset $D_g = \langle X_g, f_g \rangle$. A classifier is trained on $\hat{D}_g$ and the training algorithm selects a hypothesis $h \in H$. Its actual risk $\epsilon_r(h, f_r)$ is bounded by*

$$\epsilon_r(h, f_r) \leq \epsilon_g(h, f_g) + \lambda + \frac{1}{2} d_{h \Delta h^*}(X_g, X_r), \tag{16}$$

*where $h^*$ and $\lambda$ are defined in Eq. (8) and Eq. (9), respectively.*

### 4.4 Estimating GAN-to-Real and Domain Adaptation Bounds

Generally, it not feasible to calculate the actual values of Eq. (11) and Eq. (16) due to the probability density functions of $X_r$ and $X_g$. For our G2R bound, we propose a practical method to estimate its values. To estimate the values of the DA bound and compare in the experiments, we also revise this method for the DA bound. Our proposed method is highlighted in the following, whereas the detailed derivations are documented in the Supplementary Materials S1.2.

**G2R bound** We respectively denote the estimators of $\lambda$ and $\frac{1}{2} d_{h \Delta h^*}(X_g, X_r)$ as $\bar{\lambda}$ and $\bar{d}_{G2R}$, which are defined by

$$\bar{\lambda} = \tilde{\epsilon}_r(\hat{h}^*, f_r) + \tilde{\epsilon}_g(\hat{h}^*, f_g) \text{ and } \bar{d}_{G2R} = |\tilde{\epsilon}_g(h, \hat{h}^*) - \tilde{\epsilon}_r(h, \hat{h}^*)|, \tag{17}$$

where $\hat{h}^*$ is the hypothesis whose definition is

$$\hat{h}^* = \arg\min_{h' \in H} \hat{\epsilon}_r(h', f_r) + \hat{\epsilon}_g(h', f_g), \tag{18}$$

the notations $\tilde{\epsilon}_r(\hat{h}^*, f_r)$ and $\tilde{\epsilon}_g(\hat{h}^*, f_g)$ are testing errors mentioned in Sec.3.3, and the notation $\tilde{\epsilon}_g(h, \hat{h}^*)$ represents the expected disagreement between $h$ and $\hat{h}^*$ evaluated on $\tilde{X}_g$, i.e., $\tilde{\epsilon}_g(h, \hat{h}^*) = \mathbb{E}_{x \sim \tilde{X}_g}\left[\mathbb{I}[h(x) \neq \hat{h}^*(x)]\right]$. We denote the estimator of our G2R bound as $\bar{B}_{G2R}$, which is defined by

$$\bar{B}_{G2R} = \tilde{\epsilon}_g(h, f_g) + \bar{\lambda} + \bar{d}_{G2R}. \tag{19}$$

**DA bound** We denote the estimator of $\frac{1}{2} d_{H \Delta H}(X_g, X_r)$ as $\bar{d}_{DA}$, which is defined by

$$\bar{d}_{DA} = |1 - 2\tilde{\psi}_{rg}(h_{DA})|, \tag{20}$$

where $h_{DA}$ is the hypothesis whose definition is

$$h_{DA} = \arg\max_{h' \in H} |1 - 2\hat{\psi}_{rg}(h')|, \tag{21}$$

and the terms $\hat{\psi}_{rg}(h')$ as well as $\tilde{\psi}_{rg}(h')$ are defined by

$$\begin{aligned} \hat{\psi}_{rg}(h') &= \frac{1}{2} \left( \mathbb{P}_{x \sim \hat{X}_g}[h'(x) = 0] + \mathbb{P}_{x \sim \hat{X}_r}[h'(x) = 1] \right), \text{ and} \\ \tilde{\psi}_{rg}(h') &= \frac{1}{2} \left( \mathbb{P}_{x \sim \tilde{X}_g}[h'(x) = 0] + \mathbb{P}_{x \sim \tilde{X}_r}[h'(x) = 1] \right). \end{aligned} \tag{22}$$

Accordingly, we denote the estimator of the DA bound as $\bar{B}_{DA}$, which is defined by

$$\bar{B}_{DA} = \tilde{\epsilon}_g(h, f_g) + \bar{\lambda} + \bar{d}_{DA}. \tag{23}$$

## 5 Experimental Evaluation

In this section, according to our proposed G2R bound, we first identify the two factors influencing the generalization gap in our considered scenario. By considering the identified two factors as the control variables in our experimental evaluation, we conduct experiments to show the relationships between the real performance of trained classifiers, $\bar{d}_{G2R}$, $\bar{d}_{DA}$, $\bar{B}_{G2R}$, and $\bar{B}_{DA}$. Please note that our experiments aim to show that our proposed G2R bound and devised estimation method from the prior section can be employed on real-world problems, and do not aim for producing the best classification accuracy.

### 5.1 Factors Influencing the Generalization Gap

Based on Eq. (16), we can identify the two terms that affect the generalization gap between $\epsilon_g(h, f_g)$ and $\epsilon_r(h, f_r)$: $d_{h\Delta h^*}(X_g, X_r)$ and $\lambda$. After further examining these two terms, we find that the two root factors are 1) the discrepancy between $X_r$ and $X_g$ and 2) the inconsistency between $f_r$ and $f_g$. It is clear that the discrepancy between $X_r$ and $X_g$ contributes to $d_{h\Delta h^*}(X_g, X_r)$ since $d_{h\Delta h^*}(X_g, X_r)$ is a pseudo-metric and thus satisfies $d_{h\Delta h^*}(X_g, X_r) > 0 \Rightarrow X_g \neq X_r$. The relation between $\lambda$ and the inconsistency between $f_r$ and $f_g$ is not obvious. Thus, we utilize an example to illustrate their relation in the following. We assume a real image $x_r$ is labeled as 0 by $f_r$, whereas a synthetic image $x_g$ similar to $x_r$ is labeled as 1 by $f_g$. Under this assumption, a hypothesis $h$ classifying $x_r$ as 0 can reduce the value of $\epsilon_r(h, f_r)$, but $h$ may classify $x_g$ as 0 and hence increases the value of $\epsilon_g(h, f_g)$. As a result, the inconsistency between $f_r$ and $f_g$ affects $\lambda$ based on Eq. (9).

### 5.2 Experiment Setting

After having identified the two key factors, we conjecture that the generalization gap can be reduced if the discrepancy between $X_g$ and $X_r$ is reduced, or the inconsistency between $f_g$ and $f_r$ is reduced. This can be done if we optimize Eq. (4) well because of the following two reasons. First, $L_s$ can reduce the discrepancy between $X_r$ and $X_g$, thereby reducing $d_{h\Delta h^*}(X_g, X_r)$. Second, $L_c$ can decrease the inconsistency between $f_r$ and $f_g$ and hence decreases $\lambda$. Based on these two reasons, we conduct two sets of experiments to justify our conjecture. In both sets of experiments, we also compare our method with the prior work of [4] to demonstrate that our proven bound does not suffer from over-estimation. *1st Experiments – Discrepancy between $X_r$ and $X_g$*: In this set of experiments, we used the AC-GAN models on different training stages to examine the relationship between the generalization capability of classifiers and the discrepancy between $X_r$ and $X_g$. We trained an AC-GAN model with $500,000$ iterations with the hyper-parameter $\alpha = 1$ and selected several checkpoint models ranging from the earlier iterations to later iterations during the training. Afterwards, based on the selected models, we generated their corresponding synthetic datasets and trained several classifiers on such synthetic datasets. *2nd Experiments – Inconsistency between $f_r$ and $f_g$*: In this set of experiments, we altered the values of $\alpha$ in Eq. (4) to control the weights of $L_c$ and to investigate how the inconsistency between $f_r$ and $f_g$ influences the generalization capability of classifiers. We trained several AC-GAN models with different $\alpha$ values ranging from $0.01$ to $3$. After the models trained with $300,000$ iterations, we used the models to synthesize the related datasets, and then trained several classifiers on such datasets. In all of our experiments, we used PG-GANs [14], the state-of-the-art architecture of AC-GANs, as our AC-GAN models and LeNet [17] as our classifiers. Our experiments were evaluated on three datasets including MNIST [18], SVNH [24], and CIFAR10 [15]. Other implementation details are depicted in the Supplementary Materials S1.3.

### 5.3 Experimental Results and Discussion

Fig.1 presents some representative samples from datasets $D_g$ and $D_r$. Besides showing $\tilde{\epsilon}_r(h, f_r)$ and $\tilde{\epsilon}_g(h, f_g)$, Fig.2 utilizes our estimation method in Sec.4.4 to report $\bar{\lambda}$, $\bar{d}_{G2R}$, $\bar{d}_{DA}$, $\bar{B}_{G2R}$ and $\bar{B}_{DA}$. For 1st experiments, comparing the top three rows with the last row in Fig.1, we observe that the discrepancy of $X_r$ and $X_g$ decreases from the appearance viewpoint when the training iteration increases. Besides, the top of Fig.2 shows $\bar{d}_{G2R}$ and $\bar{B}_{G2R}$ drop when the training iteration increases, and also reveals that the trend of $\bar{d}_{G2R}$ and $\bar{B}_{G2R}$ is proportional to the trend of $\tilde{\epsilon}_r(h, f_r)$. Moreover, the top of Fig.2 demonstrates that $\bar{B}_{G2R}$ is a robust generalization bound since $\tilde{\epsilon}_r(h, f_r)$ is never larger than $\bar{B}_{G2R}$. However, $\bar{B}_{DA}$ (the prior work of [4]) cannot catch the trend of $\tilde{\epsilon}_r(h, f_r)$ since $\bar{d}_{DA}$ overestimates the distance between $X_r$ and $X_g$. For 2nd experiments, comparing the fourth, fifth and sixth rows with the last row in Fig.1, we observe that the larger $\alpha$ becomes, the smaller the inconsistency between $f_r$ and $f_g$ is. Furthermore, the bottom of Fig.2 exhibits that $\bar{\lambda}$ decreases when $\alpha$ increases, the trend of $\bar{B}_{G2R}$ and $\tilde{\epsilon}_r(h, f_r)$ is consistent, and $\bar{d}_{G2R}$ is not correlated to $\tilde{\epsilon}_r(h, f_r)$ whose decreasing trend results from the reduction of $\bar{\lambda}$. In addition, $\bar{B}_{G2R}$ is also a robust generalization bound in these experiments. However, as shown in the bottom of Fig.2, the trend of $\bar{B}_{DA}$ does not always reflect the trend of $\tilde{\epsilon}_r(h, f_r)$, and $\bar{B}_{G2R}$ is much closer to $\tilde{\epsilon}_r(h, f_r)$ than $\bar{B}_{DA}$ owing to the over-estimation of $\bar{d}_{DA}$. Based on the aforementioned observations, we summarize the experimental evaluation as follows. First, when the discrepancy between $X_r$ and $X_g$ reduces, $\bar{d}_{G2R}$ decreases, thereby improving the generalization capability, indicating by the decrease in $\tilde{\epsilon}_r(h, f_r)$. Second, when the inconsistency between $f_r$ and $f_g$ diminishes $\bar{\lambda}$ decreases, thereby improving the generalization capability. Third, we demonstrate that $\bar{B}_{G2R}$ is robust enough to guarantee the generalization capability. Last but not least, for our considered scenario, our proposed G2R bound $\bar{B}_{G2R}$ is a better measurement of generalization capability, whereas the DA bound $\bar{B}_{DA}$ [4] is not suitable here since $\bar{d}_{DA}$ is prone to overestimate the distance between $X_r$ and $X_g$.

## 6 Conclusion

To the best of our knowledge, this is the first work to propose a generalization bound for supervised learning from GAN-synthetic data. Our proposed G2R bound is a much better measurement for our considered scenario than adapting the DA bound from [4]. Based on our G2B bound, we identify the two crucial factors influencing the generalization gap and conduct two kinds of experiments to demonstrate the applicability of our G2R bound to real-world problems. Our G2R bound makes it possible to theoretically analyze the factors of the generalization gap, thereby providing the hints to improve the generalization of classifiers. As future work, we will investigate other methods to measure the distance between $X_r$ and $X_g$ without depending on $h$.

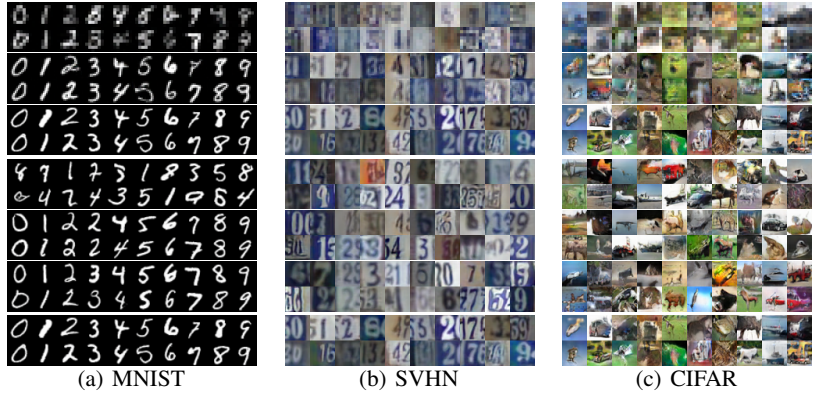

Figure 1: Sample images from the datasets $D_g$ and $D_r$ in our experiments. Each column represents a specific class of the given dataset. The top three rows show the samples from $D_g$ in 1$^{\text{st}}$ experiments when the numbers of iterations are $50,000$, $90,000$, and $500,000$. The fourth to sixth rows show the samples from $D_g$ in 2$^{\text{nd}}$ experiments when $\alpha$ is set to $0.01$, $0.03$ and $3$. The last row shows the samples from $D_r$. More example images are shown in the Supplementary Materials S1.4.

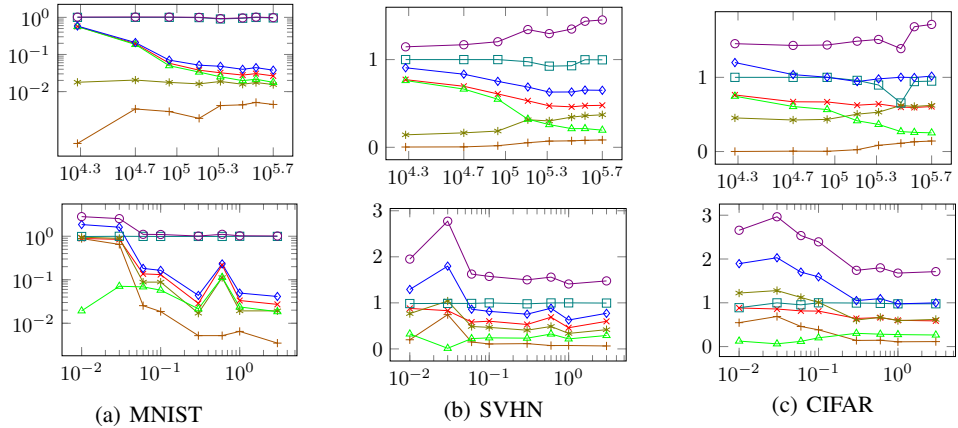

Figure 2: Quantitative results of the experiments. Each y-axis of the top and bottom parts is the error rate. The results of 1$^{\text{st}}$ experiments are shown in the top part whose x-axis is the number of training iterations. The results of 2$^{\text{nd}}$ experiments are shown in the bottom part whose x-axis is the value of $\alpha$. The legends of the curves are: $\times$ $\tilde{\epsilon}_r(h, f_r)$, $+$ $\tilde{\epsilon}_g(h, f_g)$, $*$ $\bar{\lambda}$, $\triangle$ $\bar{d}_{G2R}$, $\square$ $\bar{d}_{DA}$, $\diamond$ $\bar{B}_{G2R}$, $\circ$ $\bar{B}_{DA}$.

## S1 Supplementary Materials

### S1.1 Proof of Theorem 4.1

*Given a real dataset $D_r = \langle X_r, f_r \rangle$, a cGAN is trained on $\hat{D}_r$ based on the steps described in Sec.3.3, this cGAN generates a synthetic dataset $D_g = \langle X_g, f_g \rangle$. A classifier is trained on $\hat{D}_g$ and the training algorithm selects a hypothesis $h \in H$. Its actual risk $\epsilon_r(h, f_r)$ is bounded by*

$$\epsilon_r(h, f_r) \leq \epsilon_g(h, f_g) + \lambda + \frac{1}{2} d_{h \Delta h^*}(X_g, X_r), \tag{S1}$$

*where $h^*$ and $\lambda$ are defined in Eq. (8) and Eq. (9), respectively.*

*Proof.*

$$\begin{aligned}
\epsilon_r(h, f_r) &= \epsilon_r(h, f_r) + \epsilon_g(h, h^*) - \epsilon_g(h, h^*) + \epsilon_r(h, h^*) - \epsilon_r(h, h^*) - \epsilon_g(h, f_g) + \epsilon_g(h, f_g) \\
&\leq \epsilon_g(h, f_g) + |\epsilon_r(h, f_r) - \epsilon_r(h, h^*)| + |\epsilon_g(h, h^*) - \epsilon_g(h, f_g)| + |\epsilon_r(h, h^*) - \epsilon_g(h, h^*)| \\
&= \epsilon_g(h, f_g) + \epsilon_r(h^*, f_r) + \epsilon_g(h^*, f_g) + |\epsilon_r(h, h^*) - \epsilon_g(h, h^*)| \\
&= \epsilon_g(h, f_g) + \lambda + \frac{1}{2} d_{h \Delta h^*}(X_g, X_r).
\end{aligned} \tag{S2}$$

The notation $\epsilon_g(h, h^*)$ represents the expected disagreement between $h$ and $h^*$ evaluated on $X_g$, i.e., $\epsilon_g(h, h^*) = \mathbb{E}_{x \sim X_g}[\mathbb{I}[h(x) \neq h^*(x)]]$. The first and second lines of this proof are straightforward. For the third line, if we would like to derive the term $\epsilon_r(h^*, f_r)$, we have to know the following three facts: linearity of expected value, the triangle inequality $\mathbb{I}[a \neq b] \geq \mathbb{I}[a \neq c] - \mathbb{I}[b \neq c]$, and $\mathbb{I}[a \neq b] \geq 0, \forall a, b, c \in \mathbb{R}$. Based on these three facts, we can derive the following inequality:

$$\begin{aligned}
|\epsilon_r(h, f_r) - \epsilon_r(h, h^*)| &= \left| \mathbb{E}_{x \sim X_r}[\mathbb{I}[h(x) \neq f_r(x)]] - \mathbb{E}_{x \sim X_r}[\mathbb{I}[h(x) \neq h^*(x)]] \right| \\
&= \left| \mathbb{E}_{x \sim X_r}[\mathbb{I}[h(x) \neq f_r(x)] - \mathbb{I}[h(x) \neq h^*(x)]] \right| \\
&\leq \mathbb{E}_{x \sim X_r}[\mathbb{I}[h^*(x) \neq f_r(x)]] \\
&= \epsilon_r(h^*, f_r).
\end{aligned} \tag{S3}$$

Following the same derivation used above, the term $\epsilon_g(h^*, f_g)$ can be derived by the following inequality:

$$\left| \epsilon_g(h, h^*) - \epsilon_g(h, f_g) \right| \leq \epsilon_g(h^*, f_g). \tag{S4}$$

The last line is derived from the definition of $\lambda$ in Eq. (9), and $d_{h \Delta h^*}(X_g, X_r)$ can be written as:

$$\begin{aligned}
d_{h \Delta h^*}(X_g, X_r) &= 2 \left| \mathbb{P}_{x \sim X_g}[h(x) \neq h^*(x)] - \mathbb{P}_{x \sim X_r}[h(x) \neq h^*(x)] \right| \\
&= 2 \left| \mathbb{E}_{x \sim X_g}[\mathbb{I}[h(x) \neq h^*(x)]] - \mathbb{E}_{x \sim X_r}[\mathbb{I}[h(x) \neq h^*(x)]] \right| \\
&= 2 |\epsilon_g(h, h^*) - \epsilon_r(h, h^*)|.
\end{aligned} \tag{S5}$$

□

### S1.2 Derivation of Estimating GAN-to-Real and Domain Adaptation Bounds

#### S1.2.1 G2R bound

To estimate the terms involving $h^*$, we first search for a hypothesis $\hat{h}^* \in H$ whose definition is

$$\hat{h}^* = \arg\min_{h' \in H} \hat{\epsilon}_r(h', f_r) + \hat{\epsilon}_g(h', f_g). \tag{S6}$$

After replacing $h^*$ with $\hat{h}^*$ in our G2R bound, the revised bound is still valid since the following inequality is held:

$$\lambda = \epsilon_r(h^*, f_r) + \epsilon_g(h^*, f_g) \leq \epsilon_r(\hat{h}^*, f_r) + \epsilon_g(\hat{h}^*, f_g). \tag{S7}$$

However, the values of the actual risks in Eq. (S7) involve unknown probability density functions of $X_r$ and unsolvable integrals with probability density function of $X_g$. Fortunately, according to Sec. 2.2.3 of [1], the value of the actual risk can be estimated from the value of testing error, and then the following inequality is valid with probability $1 - \delta$:

$$\epsilon(h, f) \leq \tilde{\epsilon}(h, f) + \sqrt{\frac{1}{2m} \log \frac{2}{\delta}}, \tag{S8}$$

where $\epsilon(h, f)$ is the actual risk, $\tilde{\epsilon}(h, f)$ is the testing error of $h$, and $m$ is the number of testing samples. When $m$ is large, the rightmost square root is presumed negligible, which leads to $\tilde{\epsilon}(h, f) \approx \epsilon(h, f)$. Therefore, we can use the testing errors $\tilde{\epsilon}_g(h, f_g)$, $\tilde{\epsilon}_r(\hat{h}^*, f_r)$, and $\tilde{\epsilon}_g(\hat{h}^*, f_g)$ to estimate the actual risks $\epsilon_g(h, f_g)$, $\epsilon_r(\hat{h}^*, f_r)$ and $\epsilon_g(\hat{h}^*, f_g)$. Our proposed distance $d_{h \Delta h^*}(X_g, X_r)$ can be derived from the difference of the two actual risks between $h$ and $h^*$ given in Eq. (S5). Hence, we have:

$$d_{h \Delta h^*}(X_g, X_r) = 2|\epsilon_g(h, h^*) - \epsilon_r(h, h^*)|. \tag{S9}$$

By combining the above-mentioned equations and estimators, the estimators of $\lambda$ and $\frac{1}{2} d_{h \Delta h^*}(X_g, X_r)$, denoted as $\bar{\lambda}$ and $\bar{d}_{G2R}$ respectively, are

$$\begin{aligned} \bar{\lambda} &= \tilde{\epsilon}_r(\hat{h}^*, f_r) + \tilde{\epsilon}_g(\hat{h}^*, f_g) \text{ and} \\ \bar{d}_{G2R} &= |\tilde{\epsilon}_g(h, \hat{h}^*) - \tilde{\epsilon}_r(h, \hat{h}^*)|. \end{aligned} \tag{S10}$$

We then denote the estimator of our G2R bound as $\bar{B}_{G2R}$, which is defined by

$$\bar{B}_{G2R} = \tilde{\epsilon}_g(h, f_g) + \bar{\lambda} + \bar{d}_{G2R}. \tag{S11}$$

### S1.2.2 DA bound

The estimator $\bar{\lambda}$ in the G2R bound can also act as an estimator of $\lambda$ in DA bound. However, we need to derive a new estimator of $d_{H \Delta H}(X_g, X_r)$. First, it can be derived from a classification error via

$$\begin{aligned} d_{H \Delta H}(X_g, X_r) &= 2 \sup_{g \in H \Delta H} \left| \mathbb{P}_{x \sim X_g}[g(x) = 1] - \mathbb{P}_{x \sim X_r}[g(x) = 1] \right| \\ &= 2 \sup_{g \in H \Delta H} \left| 1 - (\mathbb{P}_{x \sim X_g}[g(x) = 0] + \mathbb{P}_{x \sim X_r}[g(x) = 1]) \right| \\ &= 2 |1 - \inf_{h' \in H \Delta H} 2\psi_{rg}(h')|, \end{aligned} \tag{S12}$$

where $\psi_{rg}(h')$ is the error that a hypothesis $h'$ tries to minimize and defined by

$$\psi_{rg}(h') = \frac{1}{2} \left( \mathbb{P}_{x \sim X_g}[h'(x) = 0] + \mathbb{P}_{x \sim X_r}[h'(x) = 1] \right). \tag{S13}$$

The term $\psi_{rg}(h')$ is assumed to be less than $0.5$. However, we cannot find such $h'$ since the value of $\psi_{rg}(h')$ involves an unknown probability density function of $X_r$ and an unsolvable integral with probability density function with $X_g$. To estimate its value, we define $\hat{\psi}_{rg}(h')$ and $\tilde{\psi}_{rg}(h')$ by:

$$\begin{aligned} \hat{\psi}_{rg}(h') &= \frac{1}{2} \left( \mathbb{P}_{x \sim \hat{X}_g}[h'(x) = 0] + \mathbb{P}_{x \sim \hat{X}_r}[h'(x) = 1] \right) \text{ and} \\ \tilde{\psi}_{rg}(h') &= \frac{1}{2} \left( \mathbb{P}_{x \sim \tilde{X}_g}[h'(x) = 0] + \mathbb{P}_{x \sim \tilde{X}_r}[h'(x) = 1] \right). \end{aligned} \tag{S14}$$

By assuming $H \subset H \Delta H$, we then search a hypothesis $h_{DA} \in H$ such that

$$h_{DA} = \arg\max_{h' \in H} |1 - 2\hat{\psi}_{rg}(h')|. \tag{S15}$$

We denote the estimator of $\frac{1}{2} d_{H \Delta H}(X_g, X_r)$ as $\bar{d}_{DA}$, which is defined by

$$\bar{d}_{DA} = |1 - 2\tilde{\psi}_{rg}(h_{DA})|. \tag{S16}$$

In order to demonstrate the over-estimation of $d_{H \Delta H}(X_g, X_r)$, we intend to find a lower bound of $d_{H \Delta H}(X_g, X_r)$ such that this lower bound still overestimates the distance between $X_r$ and $X_g$. We demonstrate that $\bar{d}_{DA}$ is a lower bound of $\frac{1}{2}d_{H\Delta H}(X_g, X_r)$ by the following derivations:

$$\begin{aligned}
\bar{d}_{DA} &= |1 - 2\tilde{\psi}_{rg}(h_{DA})| \\
&\approx |1 - 2\psi_{rg}(h_{DA})| \\
&\leq |1 - \inf_{h' \in H\Delta H} 2\psi_{rg}(h')| \\
&= \frac{1}{2}d_{H\Delta H}(X_g, X_r) \,.
\end{aligned} \quad \text{(S17)}$$

The first derivation is valid because $\tilde{\psi}_{rg}(h') \approx \psi_{rg}(h')$, which is held in Eq. (S8) when $m$ is large enough. By combining the estimators mentioned above, we denote the estimator of the DA bound as $\bar{B}_{DA}$, which is defined by

$$\bar{B}_{DA} = \tilde{\epsilon}_g(h, f_g) + \bar{\lambda} + \bar{d}_{DA} \,. \quad \text{(S18)}$$

### S1.3 Implementation Details

In all experiments, we use the public available implementation of PG-GANs[4] to generate the synthetic datasets and LeNet[5] as the classifiers. We use $\hat{D}_r$ with $n = 50,000$ to train the PG-GAN models and $\hat{D}_g$ with $n = 50,000$ to train the classifiers. We use $\tilde{D}_r$ and $\tilde{D}_g$ with $m = 10,000$ as our testing datasets. All the real datasets $D_r$ contain 10 classes. Thus, our synthetic datasets $D_g$ also contain 10 classes, and each class contains the same numbers of training and testing images (i.e., 5,000 images for training and 1,000 images for testing). All the images are resized into $32 \times 32$ px resolution with RGB channels. We use Adam optimizer as the optimizer to train PG-GAN and LeNet models. The hyper-parameters of Adam optimizer are $\beta_1 = 0, \beta_2 = 0.99$ for PG-GAN models and $\beta_1 = 0.5, \beta_2 = 0.999$ for LeNet classifiers. The learning rates are 0.001 for PG-GAN models and 0.0002 for LeNet classifiers. The batch sizes are 32 for PG-GAN models and 128 for LeNet classifiers. The training iterations for LeNet classifiers are 50,000. To obtain the hypothesis $\hat{h}_*$, we train another LeNet classifier to minimize the combined error in Eq. (S6). To obtain the hypothesis $h_{DA}$, we train another LeNet classifier to minimize the $\hat{\psi}_{rg}(h')$ in Eq. (S15). These two additional LeNet classifiers are trained under the same hyper-parameter settings as the settings of the other LeNet classifiers.

### S1.4 Sample Images from the GAN-Synthetic Datasets and the Real Datasets

The sample images from real datasets are in Fig.S1, and the sample images from GAN-synthetic datasets are in Fig.S2 and Fig.S3.

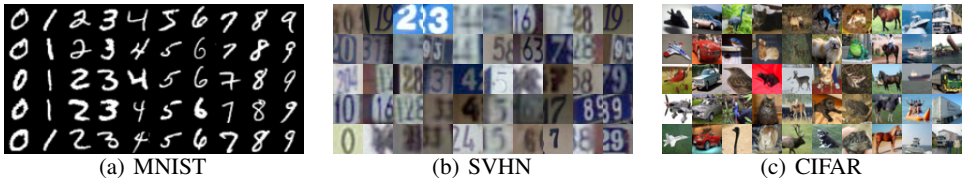

(a) MNIST        (b) SVHN        (c) CIFAR

Figure S1: Sample images from dataset $D_r$.

---

[4]https://github.com/tkarras/progressive_growing_of_gans
[5]https://github.com/activatedgeek/LeNet-5

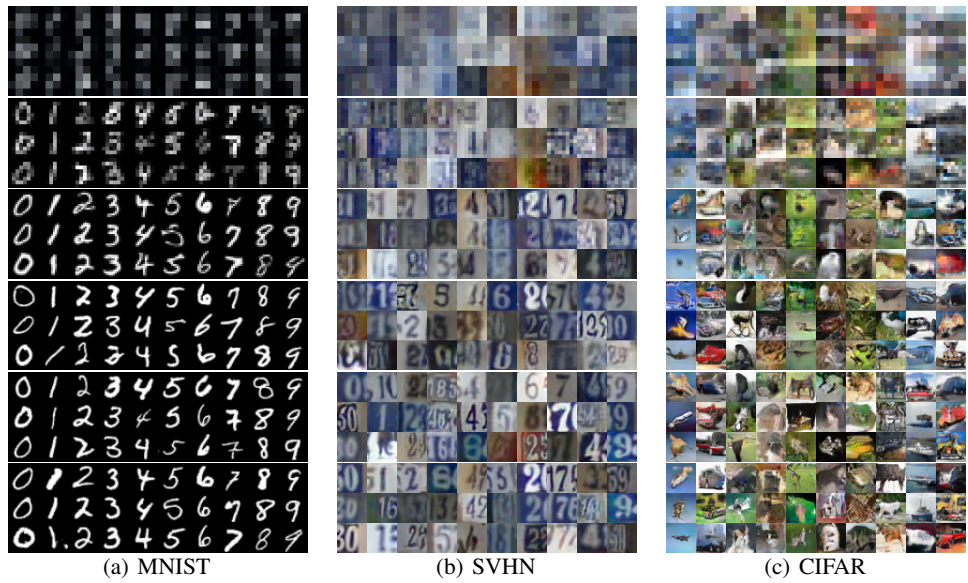

Figure S2: Sample images from GAN-synthetic datasets in 1$^{\text{st}}$ experiments when the numbers of training iterations are $20,000$, $50,000$, $90,000$, $150,000$, $300,000$ and $500,000$.

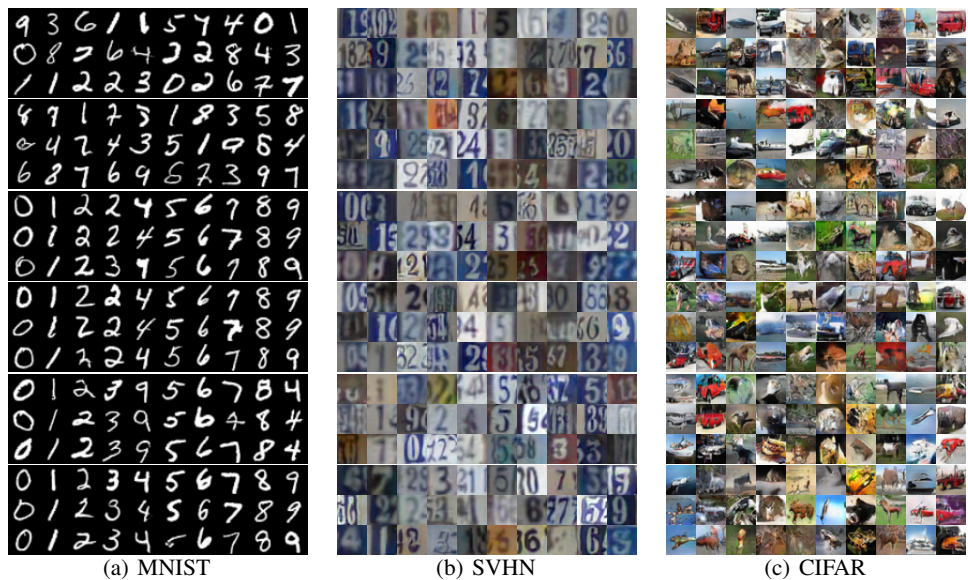

Figure S3: Sample images from GAN-synthetic datasets in 2$^{\text{nd}}$ experiments when the values of $\alpha$ are $0.01$, $0.03$, $0.06$, $0.1$, $0.6$ and $3$. Each column represents a specific class of the given dataset.